\documentclass[letterpaper, 10 pt, conference]{ieeeconf}  

\IEEEoverridecommandlockouts                              

\usepackage{amssymb}
\usepackage{amsmath}
\usepackage{array}
\usepackage{graphicx}
\usepackage{acro}
\usepackage{caption}
\usepackage{makecell}
\usepackage{pifont}
\usepackage{arydshln}
\usepackage{hyperref}
\usepackage{url}

\DeclareAcronym{ros}{short = ROS, long = Robot Operating System}
\DeclareAcronym{vlm}{short = VLM, long = Vision-Language Model}
\DeclareAcronym{orfd}{short = ORFD, long = Off-Road Freespace Detection}
\DeclareAcronym{slam}{short = SLAM, long = Simultaneous Localisation and Mapping}
\DeclareAcronym{nerf}{short = NeRF, long = Neural Radiance Field}
\DeclareAcronym{pca}{short = PCA, long = Principal Component Analysis}
\DeclareAcronym{mlp}{short = MLP, long = Multi-Layer Perceptron}

\title{\LARGE \bf
OTAS: Open-vocabulary Token Alignment for Outdoor Segmentation*
}

\author{Simon Schwaiger$^{1,2}$, Stefan Thalhammer$^{2}$, Wilfried Wöber$^{2,3}$ and Gerald Steinbauer-Wagner$^{1}$
\thanks{*This work was partly supported by the city of Vienna (MA23 – Economic Affairs, Labour and Statistics) through the project Stadt Wien Kompetenzteam für Drohnentechnik in der Fachhochschulausbildung (DrohnFH, MA23 project 35-02).}
\thanks{$^{1}$Simon Schwaiger and Gerald Steinbauer-Wagner are with Graz University of Technology, Faculty of Computer Science and Biomedical Engineering, Institute of Software Engineering and Artificial Intelligence, 8010 Graz, Austria}%
\thanks{$^{2}$Simon Schwaiger, Stefan Thalhammer and Wilfried Wöber are with University of Applied Sciences Technikum Wien, Faculty of Industrial Engineering, Research Group Digital Manufacturing, Automation and Robotics, 1200 Vienna, Austria {\tt\small schwaige@technikum-wien.at}}%
\thanks{$^{3}$Wilfried Wöber is with University of Natural Resources and Life Sciences, Department of Integrative Biology and Biodiversity Research, Institute for Integrative Nature Conservation Research, 1180 Vienna, Austria}%
}

\begin{document}


\makeatletter
\newcommand{\thickhline}{%
    \noalign {\ifnum 0=`}\fi \hrule height 1pt
    \futurelet \reserved@a \@xhline
}
\newcolumntype{"}{@{\hskip\tabcolsep\vrule width 1pt\hskip\tabcolsep}}
\newcolumntype{[}{@{\vrule width 1pt\hspace{6pt}}} \newcolumntype{]}{@{\hspace{6pt}\vrule width 1pt}}


\maketitle
\thispagestyle{empty}
\pagestyle{empty}

\begin{abstract}
Understanding open-world semantics is critical for robotic planning and control, particularly in unstructured outdoor environments.
Existing vision-language mapping approaches typically rely on object-centric segmentation priors, which often fail outdoors due to semantic ambiguities and indistinct class boundaries.
We propose OTAS—an Open-vocabulary Token Alignment method for outdoor Segmentation.
OTAS addresses the limitations of open-vocabulary segmentation models by extracting semantic structure directly from the output tokens of pre-trained vision models.
By clustering semantically similar structures across single and multiple views and grounding them in language, OTAS reconstructs a geometrically consistent feature field that supports open-vocabulary segmentation queries.
Our method operates in a zero-shot manner, without scene-specific fine-tuning, and achieves real-time performance of up to $\approx$17 fps.
On the Off-Road Freespace Detection dataset, OTAS yields a modest IoU improvement over fine-tuned and open-vocabulary 2D segmentation baselines.
In 3D segmentation on TartanAir, it achieves up to a 151\% relative IoU improvement compared to existing open-vocabulary mapping methods.
Real-world reconstructions further demonstrate OTAS' applicability to robotic deployment.
Code and a ROS 2 node are available at \url{https://otas-segmentation.github.io/}.
\end{abstract}

\section{Introduction}\label{sec:introduction}

\begin{figure}[t]
    \centering
    \includegraphics[width=0.95\linewidth]{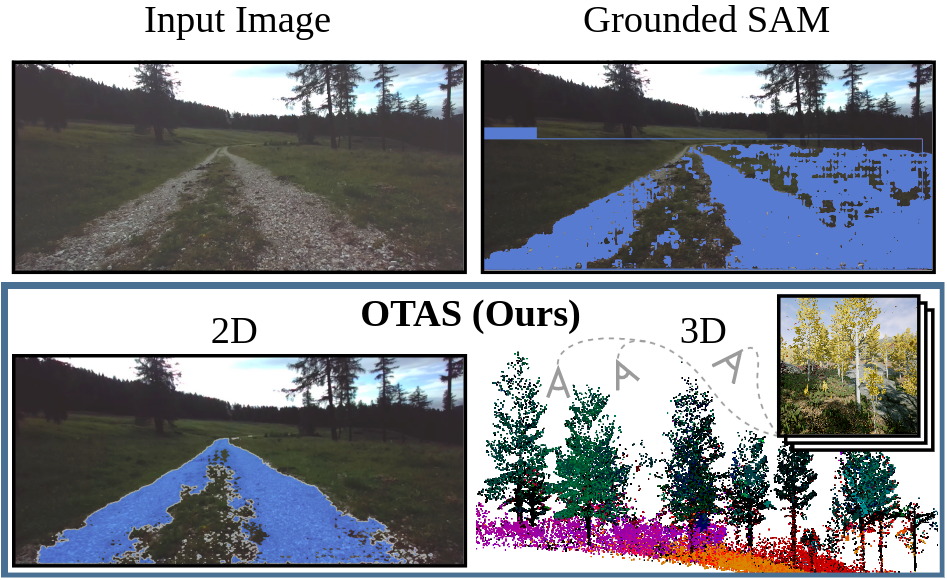}
    \caption{\textbf{OTAS} is a training-free segmentation method that aligns tokens from vision and language foundation models for robotic outdoor tasks. It operates zero-shot on single (2D) or multi-view (3D) inputs and achieves real-time operation. For 2D, the prompt \textit{"gravel road"} was used; 3D visualises \textit{"trees"} in green, \textit{"shrubbery"} in purple, \textit{"grass"} in orange, and \textit{"stone"} in red. 
    }
    \label{fig:teaser}
\end{figure}

Understanding the open world through semantics is a key challenge for robotics.
\acp{vlm}, that ground vision in language, have recently been shown to effectively provide semantics for mapping to facilitate task planning and navigation \cite{Huang2023VisualLanguageMaps, Chen2023OpenVocabulary}.
However, open-vocabulary semantic mapping methods \cite{Gu2024conceptgraphs, Yamazaki2024OpenFusion, Qiu2024Featuresplatting} rely on segmentation priors from general-purpose models to reason about the environment.
These models are trained for object-centric knowledge retrieval, therefore, they are effective for segmenting structured settings with salient objects. 
However, segmentation fails in unstructured outdoor environments, such as forests or off-road paths (see Fig.~\ref{fig:teaser}).
Unstructured, texture-rich classes relevant to outdoor robotics, such as roads and grass, are underrepresented in typical open-vocabulary image-text pair-based datasets and are often inconsistently labelled.
Visual ambiguities and indistinct boundaries, such as overlaps between gravel and grass, further complicate the task for segmentation models, which leads to imprecise segmentation masks.

In order to obtain robust semantic segmentation in unstructured outdoor environments, we introduce \textbf{OTAS}, an \textbf{O}pen-vocabulary \textbf{T}oken \textbf{A}lignment Method for Outdoor \textbf{S}egmentation.
\textit{token alignment} refers to clustering self-supervised visual tokens into coarse semantic structures, then pooling co-located \ac{vlm} tokens over these clusters for regularisation and language-grounding.

\begin{table*}[ht]
    \centering
    \vspace{1mm}
    \caption{\textbf{Comparison of Semantic Reconstruction Methods.} Assuming 10~fps as real-time—typical for low-dynamic settings like forests and agriculture—only OpenFusion and OTAS meet this threshold. Only LERF and OTAS use non-object-centric language maps. OTAS uniquely supports semantic segmentation in both 2D and 3D natively.}
    \vspace{+1.5ex}
    \centering
    \begin{tabular}{c"ccccccc} \thickhline
         \textbf{Method} & \textbf{Foundation Model} & \textbf{Real-Time} & \textbf{Zero-Shot} & \textbf{3D} & \textbf{2D} & \textbf{Representation} & \textbf{Not Object-Centric} \\ \thickhline
         LERF \cite{Kerr2023LERF} & OpenCLIP \cite{Cherti2023OpenCLIP}, DINOv2 \cite{Oquab2023DINOv2} & \ding{55} & \ding{55} & \ding{51} & \ding{55} & NeRF & \ding{51} \\
         Feature Splatting\cite{Qiu2024Featuresplatting} & CLIP \cite{Radford2021CLIP}, DINOv2 \cite{Oquab2023DINOv2}, SAM \cite{Kirillov2023SegmentAnything} & \ding{55} & \ding{55} & \ding{51} & \ding{55} & Gaussian Splatting & \ding{55} \\ 
         ConceptGraphs \cite{Gu2024conceptgraphs} & OpenCLIP \cite{Cherti2023OpenCLIP}, SAM \cite{Kirillov2023SegmentAnything} & \ding{55} & \ding{51} & \ding{51} & \ding{55} & Points & \ding{55} \\
         OpenFusion \cite{Yamazaki2024OpenFusion} & SEEM \cite{Xueyan2023SEEM} & \ding{51} & \ding{51} & \ding{51} & \ding{55} & TSDF & \ding{55} \\
         \textit{OTAS (ours)} & CLIP \cite{Radford2021CLIP}, DINOv2 \cite{Oquab2023DINOv2}, SAM2 \cite{Ravi2024Sam2} & \ding{51} & \ding{51} & \ding{51} & \ding{51} & Points & \ding{51} \\ \thickhline
    \end{tabular}
    \label{tab:sota_3d}
\end{table*}

Instead of relying on language semantics for segmentation, we cluster tokens based on visual prototypes derived from self-supervised pre-trained vision models.
Language grounding is obtained through semantic and spatial alignment over token clusters, alleviating the need for linear probing or rendering.
Optionally, multiple observations can be aligned to obtain a language-embedded reconstruction with geometric consistency.
Hence, OTAS is not subject to the object-centric bias learned by general-purpose segmentation models, despite also performing zero-shot inference.
The contributions of this study are:
\begin{itemize}
    \item a \textit{training-free token alignment} that fuses self-supervised visual tokens with language embeddings, regularising \ac{vlm} features and improving non-object-class segmentation, without per-scene optimisation; and
    \item a \textit{language-grounded 3D feature field} that enables real-time mapping and open-vocabulary querying, built from aligned tokens, requiring no per-scene trained \acp{mlp}, and no differentiable rendering. 
\end{itemize}

We demonstrate token alignment for 2D and 3D segmentation as well as semantic reconstruction tasks, where it achieves real-time inference on GPU.
OTAS improves segmentation results on \ac{orfd}~\cite{Chen2022ORFD_OFF_Net} and TartanAir~\cite{Wang2020TartanAir}.
Additional experiments on robot data demonstrate the advantage of OTAS for language-embedded reconstruction of unstructured outdoors in comparison to volumetric rendering with LERF~\cite{Kerr2023LERF} and Feature Splatting~\cite{Qiu2024Featuresplatting}.
Finally, critical design decisions, including token alignment, clustering methods, number of token clusters, and backbone choice, are ablated to motivate the recommended model configurations for robotic applications.

\section{Related Work}\label{sec:related_work}

\ac{vlm}s ground vision in language by encoding a joint feature space, typically extracting one feature per image or patch \cite{Radford2021CLIP, Cherti2023OpenCLIP}. 
Many robotic tasks, however, require fine-grained spatial relationships.
This motivates mapping \ac{vlm} features to queryable semantic maps~\cite{Gu2024conceptgraphs, Yamazaki2024OpenFusion, Qiu2024Featuresplatting, Kerr2023LERF, Nur2022CLIP-Fields, Qiu2024GeFF}.

Early \ac{vlm}-based navigation approaches detect objects, extract \ac{vlm} features per instance, and ground them on 2D occupancy grids (e.g., VLMaps \cite{Huang2023VisualLanguageMaps}, VLFM \cite{Yokoyama2024VLFM}) by interpolating features spatially. 
They rely on general-purpose detection or segmentation models, which introduce an object-centric prior into feature extraction~\cite{li2023blip}.
This paradigm has been extended to 3D. OpenFusion \cite{Yamazaki2024OpenFusion} fuses SEEM~\cite{Xueyan2023SEEM} features into a 3D semantic map through \ac{slam}.
Similarly, ConceptGraphs \cite{Gu2024conceptgraphs} uses SAM~\cite{Kirillov2023SegmentAnything} masks and OpenCLIP~\cite{Cherti2023OpenCLIP} features, projected to 3D and fused via geometric and semantic similarity. 
While effective indoors, all retain object-centric biases from their segmentation models.

An alternative direction is to reconstruct language-grounded feature fields. 
Feature Splatting \cite{Qiu2024Featuresplatting} retains object priors since it uses SAM for generating segmentation masks.
LERF \cite{Kerr2023LERF} avoids object priors by extracting multiscale OpenCLIP features, yielding dense, non-object-centric feature maps refined via neural rendering. 
Both rely on geometric consistency across views. 
Using neural scene representations, such as LERF or Feature Splatting, requires rendering, resulting in slow scene-specific training and making them neither zero-shot nor real-time capable.
Similar to rendering-based methods, OpenScene \cite{Peng2023OpenScene} distils multi-view CLIP features into a sparse 3D network.
However, this requires computationally intensive scene-specific training, making the method label-free but not training-free on new scenes.

\begin{figure*}[!ht]
    \centering
    \vspace{2mm}
    \includegraphics[width=0.75\linewidth]{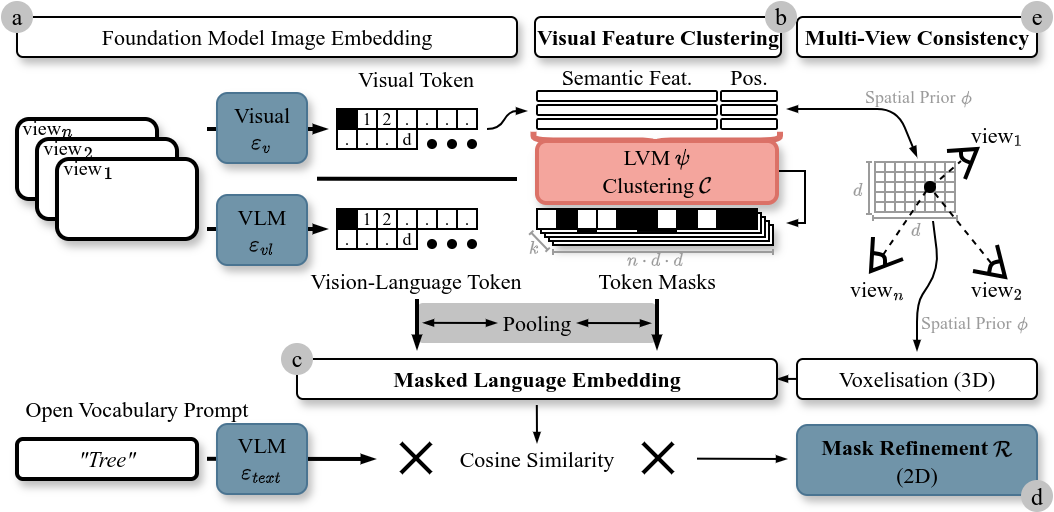}
    \caption{\textbf{Method Overview.} a) OTAS encodes input views using frozen encoders. b) Patch tokens of the visual encoder are reduced and clustered to obtain semantic masks. c) The masks are pooled with normalised patch tokens of a vision-language encoder for natural language grounding. d) A frozen mask refinement network projects semantic similarity to prompts to pixel-level. e) Clustering and pooling are optionally conditioned on environment geometry through projection.}
    \label{fig:method_overview}
\end{figure*}

Table~\ref{tab:sota_3d} compares state-of-the-art semantic reconstruction methods for outdoor robot navigation relevance. 
Key requirements include real-time performance for robot control, zero-shot applicability to new environments, and avoidance of object-centric priors for accurate segmentation of non-salient objects. 
OTAS is the only method meeting all criteria, enabling training-free and fast robotic deployment.

\section{Method}\label{sec:methods}

\acp{vlm}, such as Grounded SAM and SEEM are biased towards object-centric knowledge retrieval \cite{Zhang2025QuantifyingLimitsSAM}.
This becomes especially problematic in the unstructured environments of outdoor robotics, where the semantic classes of interest fall outside the a priori encoded object-centric knowledge.
Examples of such classes are road, woods, and shrubbery, which, however, are highly relevant to mobile outdoor robotics.

Self-supervised pre-trained vision foundation models, such as DINOv2 \cite{Oquab2023DINOv2}, do not have this limitation, since they are not trained directly on segmentation tasks.
Their training process results in an emergent semantic organisation of the feature space, where semantically similar classes are embedded adjacently.
Hence, we disentangle the open-vocabulary semantic segmentation by using DINOv2 for coarse zero-shot semantic clustering, followed by natural language grounding by pooling over CLIP's vision-language embeddings. 
Input views are embedded by the frozen vision and vision language encoders, see Figure~\ref{fig:method_overview} (a).
Output tokens of the vision encoder are clustered to obtain semantic structures (b), and aligned with vision language tokens to obtain language grounding (c).
The language-grounded semantic clusters are used as priors for optional zero-shot upscaling to pixel level (d) \cite{Ravi2024Sam2}.
Optional spatial regularisation of steps b and c increases geometric consistency and allows multi-view reconstruction and segmentation (e).

\subsection{Visual Feature Clustering}

Given a monocular input image $I \in \mathbb{R}^{H \times W \times 3}$, our goal is to generate a semantic segmentation mask guided by both vision and language.
The input image is first processed by a frozen vision encoder $\mathcal{E}_v$ to produce a coarse spatial feature map $F_v = \mathcal{E}_v(I) \in \mathbb{R}^{H' \times W' \times C_v}$.
To align vision with language, $F_v$ is interpolated to a shared feature dimension $d$ using bilinear interpolation.
The interpolated features are then flattened and L2 normalised, denoted by $\mathbf{f}_v$.
The flattened feature map is decorrelated and reduced in dimensionality using a latent variable model (LVM) $\psi$, resulting in $\hat{\mathbf{f}}_{LVM} = \psi(\hat{\mathbf{f}_v}) \in \mathbb{R}^{d \cdot d \times C_{r}}$, where the reduced feature dimension $C_{r}$ is a hyperparameter.

Subsequently, a clustering model $\mathcal{C}$ is applied to the flattened feature map $\hat{\mathbf{f}_v}$ to derive $k$ clusters, that constitute mixtures of visual tokens, referred to as visual prototypes.
The affiliation of each data point to a cluster is denoted by $\mathcal{C} = \{\mathcal{C}_1, ..., \mathcal{C}_{d \cdot d}\}, C_j \in \{1,...,k\}\forall j$, representing the assignment of the latent representations $\hat{\mathbf{f}}_{LVM}$ to a visual prototype.
The clusters are interpreted as a set of $k$ binary masks $\mathcal{M} = \{\mathcal{M}_1, ..., \mathcal{M}_k\}$, where each mask $\mathcal{M}_i \in \{0,1\}^{n \times d \times d}$ corresponds to the shared feature dimension $d$ across $n$ input images.

\subsection{Masked Language Embedding}

DINOv2 embeddings are not correlated with semantics such as language.
An intuitive way to retrieve semantic categories is linear probing.
This, however, requires annotated data in the target domain.
Instead, we use a vision-language encoder $\mathcal{E}_{vl}$ to produce language-grounded tokens and align them with the visual tokens, resulting in $F_{vl} = \mathcal{E}_{vl}(I) \in \mathbb{R}^{H_{vl} \times W_{vl} \times C_{vl}}$.
To extract dense patch-level features from the vision-language encoder, we use value features from the final attention layer rather than after global pooling, which preserves the vision-language association for dense prediction \cite{Zhou2021MaskCLIP}.
These tokens are subsequently interpolated to match $d$ using nearest neighbour interpolation ($\mathcal{U}_{nn}$): $F_{vl}^{shared} = \mathcal{U}_{nn}(F_{vl}) \in \mathbb{R}^{d \times d \times C_{vl}}$.
We adopt masked average pooling (MAP) to address token alignment, following \cite{Qiu2024Featuresplatting}, who showed its regularising effect on VLMs. Unlike prior work, we apply MAP over coarse feature maps in the shared embedding space rather than at pixel level. 
MAP computes the mean language feature vector for each mask. This is done per image, also in the case of multi-view inputs.
\begin{equation}
    F_{pooled}(x,y) = \frac{1}{|M_c|} \sum_{(x,y) \in M_c} F_{vl}^{shared}(x,y)
\end{equation}

\begin{figure}[t]
    \centering
    \includegraphics[width=0.8\linewidth]{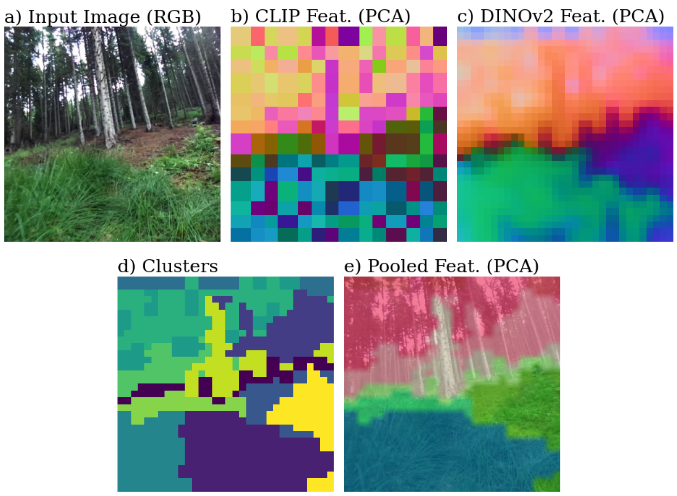}
    \caption{\textbf{Feature comparison.} CLIP (b) \cite{Radford2021CLIP} features include view-dependent noise that is detrimental to segmentation accuracy \cite{Qiu2024Featuresplatting}. We achieve regularisation in non-object-centric environments by extracting visual prototypes from DINOv2 (c) \protect\cite{Oquab2023DINOv2}, with k-Means clustering (d) and language grounding via feature pooling (e).
    }
    \label{fig:feature-types}
\end{figure}

Since each patch is only assigned to a single mask in $\mathcal{M}$, the resulting $F_{pooled}$ is a feature map of shape $d \times d \times C_{vl}$.
$F_{pooled}$ represents a language-grounded image embedding, regularised by the token mask structure (see Figure \ref{fig:feature-types}).
Ultimately, pooled features are normalised using the L2 norm.

A frozen text encoder $\mathcal{E}_{text}$ maps text prompts to the vision-language feature dimension $F_{text} = \mathcal{E}_{text}(t) \in \mathbb{R}^{C_{vl}}$. 
Cosine similarity between $F_{text}$ and each feature in $F_{pooled}$ produces a similarity map of shape $d \times d$.
As done by \cite{Kerr2023LERF, Qiu2024Featuresplatting}, $\mathcal{E}_{text}$ and the similarity computation are applied to a set of positive prompts $t^+$ and negative prompts $t^-$, indicating target and undesired concepts, respectively, resulting in the combined similarity map $\mathcal{S}_{combined}$.
\begin{equation}
    \mathcal{S}_{combined} = \sum_{ t \in t^+} \mathcal{S}(t, F_{pooled}) - \sum_{ t \in t^-} \mathcal{S}(t, F_{pooled})
\end{equation}

\subsection{Mask Refinement}

We use the similarity map as a language-grounded prior to obtain a binary pixel-level segmentation mask $M$.
Depending on the used encoders and interpolation to the shared feature resolution $d$, the similarity map resolution will be lower than the input image resolution.
Typically, the similarity map is at 1/8th or 1/16th of the input image resolution.
In order to refine the coarse mask we employ a frozen mask refinement network $\mathcal{R}$ that takes the image $I$ and the similarity map $\mathcal{S}_{combined}$ as input and outputs the final high-resolution segmentation mask. 
\begin{equation}
    M = \mathcal{R}(I, \mathcal{U}_{bl}(\mathcal{S}_{combined})) \in \{0,1\}^{H \times W}
\end{equation}

\subsection{Multi-View Consistency}

To expand OTAS to the multi-view case, information is aggregated over multiple views using the depth map $D \in \mathbb{R}^{H \times W}$ and camera pose $T \in SE(3)$ associated with each frame.
During image embedding, $D$ is projected to 3D points $P \in \mathbb{R}^{N \times 3}$.
Median depth $\tilde{D}$ is sampled in each grid of size $d \times d$ to align the 3D points with the vision and vision language features.
Using camera intrinsics $K$, 3D points $P$ are projected to the image plane via $P = \pi(\tilde{D}, K) \in \mathbb{R}^{d \cdot d \times 3}$. A mapping $\phi_i$ tracks the relationship between 3D points $P_i$ and patch indices $(i,j)$. The points are transformed to a global coordinate frame using camera poses $\{T_1, ..., T_n\}$ to construct $P_{global} = \bigcup_{i=1}^n T_i P_i$.

\textbf{Spatially Conditioned Clustering.}
The global point positions and relationship $\phi$ allow conditioning the visual feature clustering by concatenating semantic features $F_{v}^{shared}$ with 3D coordinates $P_{global}$. 
This yields a combined feature map $F_{spatial} \in \mathbb{R}^{d \cdot d \times (C_{v} + 3)}$ that replaces $F_{v}^{shared}$ as input for the LVM, where each feature vector $F_{spatial}(i,j)$ contains both semantic and spatial information for the corresponding point $p$.

\textbf{Spatially Conditioned Pooling.}
After pooling the visual and vision-language features for each input view separately, each $F_{pooled}$ is projected on the global point cloud $P_{global}$ using the relationship $\phi$, resulting in a spatial 3D feature volume $P_{semantic} \in \mathbb{R}^{d \cdot d \times (C_{vl} + 3)}$ where $P_{semantic} = \text{concat}(F_{pooled}(i,j), P_{global}(p)) \mid p \in P_{global}, (i,j) = \phi_i(p)$.
The feature volume consists of keypoint position and language-grounded feature embedding pairs.
Knowing the keypoint position, the feature volume is downsampled using a configurable voxel-size $v$.
During downsampling, all pooled features in a voxel are linearly interpolated to further condition the language-embeddings with spatial context, where $\hat{P}_{semantic} = (\frac{1}{|V_k|}\sum_{(f,p) \in V_k} f)$ with $V_k = \{(f,p) \in P_{semantic} \mid \lfloor \frac{p}{v} \rfloor = k\}$.
$\hat{P}_{semantic}$ describes a language-queryable 3D occupancy grid directly usable for robotic applications such as obstacle avoidance and goal-based navigation.

\section{Experiments}\label{sec:experiments}

\textbf{Datasets and Metrics.}
Monocular semantic segmentation is evaluated on the Off-Road Freespace Detection Dataset (ORFD) \cite{Chen2022ORFD_OFF_Net}.
ORFD aims to identify traversable road types in the outdoors, such as gravel, dirt and sand.
RELLIS-3D \cite{Jiang2021Rellis} is used in ablations as a stress test due to the high semantic overlap between annotated classes and fuzzy class boundaries.
3D feature reconstruction is evaluated on TartanAir \cite{Wang2020TartanAir}, a large-scale, photorealistic synthetic dataset for visual SLAM and robot navigation.

Since TartanAir does not provide 3D ground truth labels, 3D labels are generated for all methods by projecting 2D labels onto the reconstructed point clouds via majority vote over each point's 5 nearest neighbours, following \cite{Alama2025RayFronts}.
In order to evaluate unstructured outdoor segmentation, we evaluate segmenting vegetation, labels 152 and 109. 
Following previous work \cite{Min2024AutonomousUnstructuredReview}, Intersection over Union (IoU), F-score (Fsc), Precision (Pre), and Recall (Rec) are evaluated for all quantitative experiments.
Practical applicability to robotic applications is demonstrated through qualitative real-world reconstruction in the alps \cite{Eder2023RoboNav} and runtime and memory footprint analysis in 2D and 3D.

\textbf{Implementation Details.}
OTAS is provided in three configurations.
All models use CLIP ViT-B-16 \cite{Radford2021CLIP, Zhou2021MaskCLIP} and DINOv2 ViT-S-14 with 4 registers \cite{Oquab2023DINOv2, Darcet2023VitRegisters}.
\textit{OTAS Small} uses a shared feature dimension of $d=16$ and SAM2.1 Hiera-T \cite{Ravi2024Sam2} for mask refinement.
\textit{OTAS Large} uses $d=32$ and SAM2.1 Hiera-L.
\textit{OTAS Spatial} uses $d=64$, a voxel-size of $v=0.5m$, and no mask refinement, as segmentations are regularised geometrically. 
All models use GPU-accelerated \ac{pca} for $\psi$ and k-Means for $C$.
Evaluations are done on an Intel i7-12700 CPU and NVIDIA 4070 Ti Super GPU.

\subsection{3D Outdoor Segmentation}

\begin{table}[]
    \centering
    \vspace{1mm}
    \caption{\textbf{3D Vegetation Segmentation on TartanAir.} 
    All methods reconstruct a language-grounded point cloud given known camera poses. Sec denotes total reconstruction time excluding evaluation. We compare per point segmentation performance in identifying vegetation.}
    \begin{tabular}{lccccc}
        \thickhline
        & \multicolumn{5}{c}{\textbf{Amusement}} \\
        & IoU$\uparrow$ & Fsc$\uparrow$ & Pre$\uparrow$ & Rec$\uparrow$ & Sec$\downarrow$ \\
        OpenFusion  \cite{Yamazaki2024OpenFusion} & 23.13 & 37.09 & 39.17 & 37.86 & \underline{55} \\
        ConceptGraphs \cite{Gu2024conceptgraphs} & \underline{34.86} & \underline{46.15} & \underline{47.00} & \underline{48.17} & 2201 \\
        \textit{OTAS Spatial} (\textbf{Ours}) & \textbf{47.11} & \textbf{64.04} & \textbf{65.16} & \textbf{65.48} & \textbf{22} \\ \thickhline

        & \multicolumn{5}{c}{\textbf{Gascola}} \\
        & IoU$\uparrow$ & Fsc$\uparrow$ & Pre$\uparrow$ & Rec$\uparrow$ & Sec$\downarrow$ \\
        OpenFusion  \cite{Yamazaki2024OpenFusion} & 10.24 & 18.37 & 18.23 & 20.36 & \underline{52} \\
        ConceptGraphs \cite{Gu2024conceptgraphs} & \underline{30.68} & \underline{38.03} & \underline{30.68} & \underline{50.00} & 333 \\
        \textit{OTAS Spatial} (\textbf{Ours}) & \textbf{67.87} & \textbf{80.27} & \textbf{79.23} & \textbf{81.73} & \textbf{12} \\ \thickhline

        & \multicolumn{5}{c}{\textbf{Seasonsforest}} \\
        & IoU$\uparrow$ & Fsc$\uparrow$ & Pre$\uparrow$ & Rec$\uparrow$ & Sec$\downarrow$ \\
        OpenFusion \cite{Yamazaki2024OpenFusion} & \underline{25.09} & \underline{35.18} & 47.38 & 39.07 & \underline{53} \\
        ConceptGraphs \cite{Gu2024conceptgraphs} & 17.39 & 28.96 & \underline{51.06} & \underline{52.25} & 151 \\
        \textit{OTAS Spatial} (\textbf{Ours}) & \textbf{43.63} & \textbf{57.23} & \textbf{57.09} & \textbf{57.42} & \textbf{10} \\ \thickhline

        & \multicolumn{5}{c}{\textbf{Seasonsforest Winter}} \\
        & IoU$\uparrow$ & Fsc$\uparrow$ & Pre$\uparrow$ & Rec$\uparrow$ & Sec$\downarrow$ \\
        OpenFusion \cite{Yamazaki2024OpenFusion} & 22.16 & 36.01 & 39.37 & 40.84 & \underline{103} \\
        ConceptGraphs \cite{Gu2024conceptgraphs} & \underline{36.48} & \underline{53.33} & \underline{54.26} & \underline{54.49} & 479 \\
        \textit{OTAS Spatial} (\textbf{Ours}) & \textbf{39.61} & \textbf{55.13} & \textbf{56.22} & \textbf{55.33} & \textbf{18} \\ \thickhline
    \end{tabular}
    \label{tab:seg3d_tartanair}
\end{table}

\begin{figure*}[h]
    \centering
    \vspace{2mm}
    \includegraphics[width=0.9\linewidth]{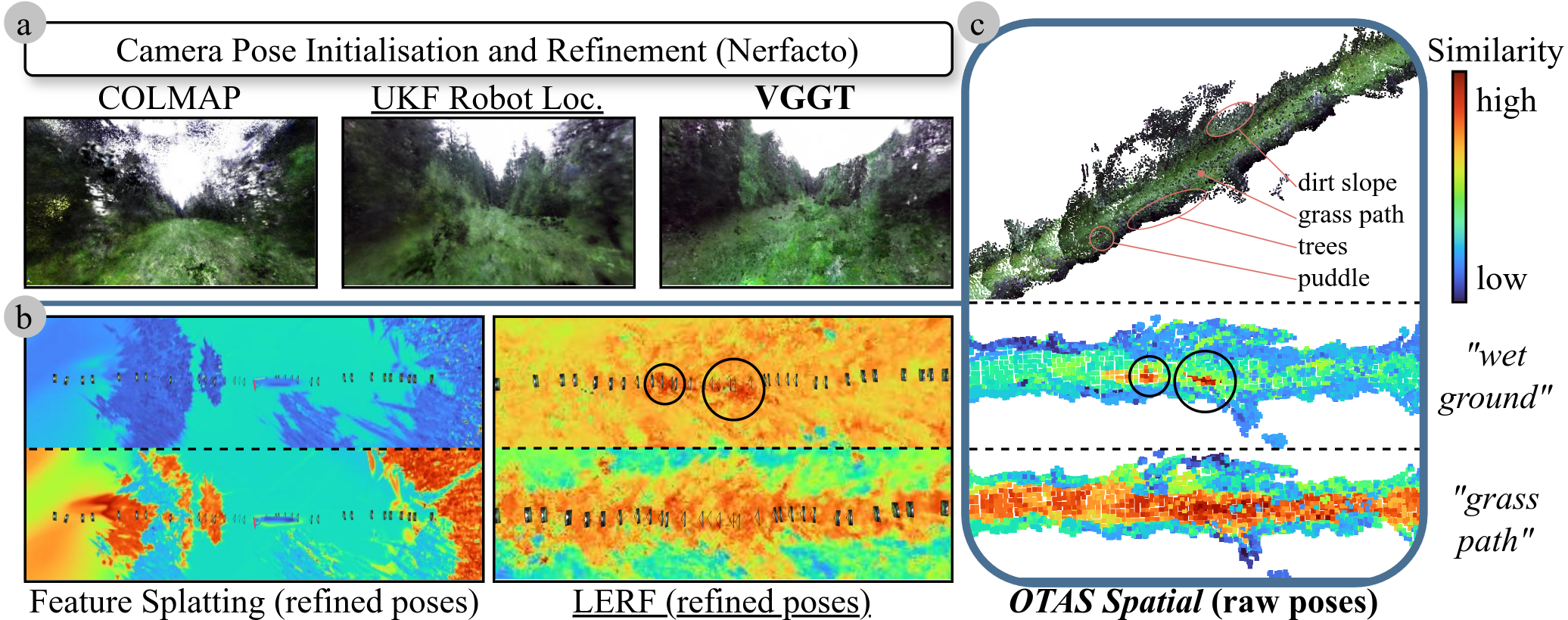}
    \caption{\textbf{Alpine Ground Analysis.} Language-embedded reconstruction requires accurate camera poses. a) Reconstruction obtained using COLMAP, UKF Robot Localisation, and VGGT. All poses are refined using Nerfacto. b) Semantic similarity of Feature Splatting and LERF to prompts. c) Semantic reconstruction and prompt similarity of \textit{OTAS Spatial}.}
    \label{fig:seg3d_robonav}
\end{figure*}

Semantic mapping is evaluated against Concept Graphs \cite{Gu2024conceptgraphs} and OpenFusion \cite{Yamazaki2024OpenFusion}, since they create language-embedded 3D pointclouds, similarly to OTAS.
Both methods serve as the state of the art for zero-shot semantic scene reconstruction in robotics, as they are not domain-specific and do not require a pretrained map prior (e.g., encoded in an \ac{mlp}).
Since both methods do not directly provide semantic labels, but rather language-grounded point clouds, we threshold using the same language queries as for OTAS.

\textbf{TartanAir}. Table~\ref{tab:seg3d_tartanair} presents 3D segmentation results on outdoor scenes of TartanAir using the first annotated trajectory. 
OTAS improves all evaluated metrics over OpenFusion and ConceptGraphs on Amusement, Gascola, and Seasonsforest. 
Especially in environments with barely any discrete objects, such as Gascola, the margin for improvement is huge, reaching up to 151$\%$ on IoU. 
The lower contrast reduces segmentation quality of object-centric open-vocabulary segmentation, highlighting the advantages of OTAS for outdoor robotics.
We observe that ConceptGraphs performs closer in snowy scenes of Seasonsforest Winter. This is likely due to the high contrast between objects and the uniform snow, which enhances object boundaries and thus benefits object-centric methods. 

\subsection{2D Outdoor Segmentation}

\begin{table}[]
    \centering
    \caption{\textbf{2D Semantic Segmentation on ORFD.}
    We include the current state of the art in fine-tuned off-road segmentation methods as well as other zero-shot segmentation methods that serve as the baseline for language-grounded semantic scene representations. The $\dagger$ indicates results optioned from the reimplementation by \cite{Jiahang2024RoadFormer}.}
    \begin{tabular}{l"cccc"r}
        \thickhline
        \textbf{Fine-tuned Methods} & IoU & Fsc  & Pre & Rec \\
         OFF-Net \cite{Chen2022ORFD_OFF_Net} & 82.30 & 90.30 & 86.60 & 94.30 \\
        RTFNet\textsuperscript{$\dagger$} \cite{Sun2019RTFNet} & 90.70 & 95.10 & 93.80 & \underline{96.50} \\
        RoadFormer \cite{Jiahang2024RoadFormer} & 92.51 & 96.11 & 95.08 & \textbf{97.17} \\ 
        M2F2-Net \cite{Hongliang2023M2F2-Net} & 93.10 & 96.40 & 97.30 & 95.50 \\
        NAIFNet \cite{Lv2024NaifNet} & \underline{94.10} & \underline{97.00} & 97.50 & 96.40 
        \\ \hline
        \textbf{Zero-Shot Methods} & IoU & Fsc  & Pre & Rec & fps \\
        SEEM \cite{Xueyan2023SEEM} & 51.31 & 59.12 & 61.44 & 60.93 & \textbf{15.0} \\ 
        Grounded SAM \cite{Ren2024GroundedSAM} & 90.49 & 94.13 & 95.12 & 93.32 & 1.8 \\ 
        Grounded SAM-2 \cite{IDEAResearchGroundedSAM2} & 93.32 & 96.38 & \underline{97.73} & 95.38 & 3.8 \\ 
        \textit{OTAS Small} (\textbf{Ours}) & 91.72 & 95.59 & 96.93 & 94.58 & \underline{11.2} \\
        \textit{OTAS Large} (\textbf{Ours}) & \textbf{94.34} & \textbf{97.05} & \textbf{97.83} & 96.39 & 5.1 \\ \thickhline
    \end{tabular}
    \label{tab:seg2d_orfd}
\end{table}

\textbf{ORFD.} This section compares OTAS to the state of the art for fine-tuned and open-vocabulary 2D semantic segmentation.
For open-vocabulary, we report Grounded SAM \cite{Ren2024GroundedSAM} and SEEM \cite{Xueyan2023SEEM}, since these are the models used by Concept Graphs \cite{Gu2024conceptgraphs} and OpenFusion \cite{Yamazaki2024OpenFusion} respectively.
SAM- and SEEM-based methods currently define the state of the art in open-vocabulary segmentation, and are therefore natural baselines.
While the original Grounded SAM-2 implementation relies on SAM2, we run it with the improved SAM2.1 Hiera-L segmentation head to provide a best-case scenario and fair comparison to our method.

\begin{table*}[!ht]
    \centering
    \vspace{1mm}
    \caption{\textbf{Influence of Model Size.} Comparison of accuracy, memory and fps of OTAS on ORFD. \textit{No Token Alignment} ablates token alignment and directly prompts from CLIP similarity maps. Both OTAS versions without mask refinement significantly outperform directly prompting mask refinement from CLIP similarity maps (line 2) w.r.t. segmentation quality and throughput, validating our token alignment strategy.}
    \begin{tabular}{l"c|c|c|c|c"c|c} \thickhline
         \textbf{Model} & Mask Refinement & IoU (\%) & Fsc (\%)  & Pre (\%) & Rec (\%) & GPU Mem. (GB) & fps (s\textsuperscript{-1}) \\ \hline
        
        No Token Alignment (GPU) & no & 68.25 & 80.46 & 79.57 & 82.48 & 1.6 & $\approx$25 \\ 
        & yes & 75.48 & 84.54 & 92.90 & 82.03 & 2.4 & $\approx$13 \\ \hdashline 

        Small (GPU) & no & 84.71 & 91.35 & 91.12 & 92.84 & 1.6 & $\approx$17 \\ 
        & yes & 91.72 & 95.59 & 96.93 & 94.58 & 2.4 & $\approx$11 \\ \hdashline 
        
        Small (CPU) & no & 84.80 & 91.41 & 91.20 & 92.87 & - & $\approx$1.6 \\
        & yes & 91.71 & 95.58 & 96.93 & 94.57 & - & $\approx$0.38 \\ \hdashline

        Large (GPU) & no & 87.02 & 92.69 & 92.3 & 94.4 & 1.6 & $\approx$15 \\
        & yes & 94.34 & 97.05 & 97.83 & 96.39 & 3.5 & $\approx$5 \\ \thickhline 
        
    \end{tabular}
    
    \label{tab:ablation_speed}
\end{table*}

Table~\ref{tab:seg2d_orfd} presents results on \ac{orfd}. 
OTAS achieves the highest IoU, Fsc and precision among fine-tuned and zero-shot methods. 
OTAS reports the highest recall among zero-shot methods.
Yet, the segmentation recall of the fine-tuned RoadFormer marginally improves over OTAS.
Interestingly, this phenomenon can be observed for all zero-shot methods. 
They exhibit lower recall as compared to fine-tuned methods. 
This is a consequence of the lack of dense supervision for specific classes and the necessity to generalise over broad, noisy semantics, whereas fine-tuned models directly optimise for segmenting the specific classes, including dataset characteristics like annotation errors and noise.

Token alignment, runtime and memory scaling as well as backbone choice are ablated in Sec.~IV-D (see Tab.~\ref{tab:ablation_speed}-\ref{tab:cluster-types} and Fig.~\ref{fig:cluster-numbers}).
These experiments demonstrate that OTAS maintains efficiency across varying cluster sizes and performs across different foundation models beyond the highlighted DINOv2/CLIP version.

\subsection{Real-World Semantic Reconstruction}

This section directly compares OTAS to LERF \cite{Kerr2023LERF} and Feature Splatting \cite{Qiu2024Featuresplatting} for semantic reconstruction in the foothills of the Alps.
While neither zero-shot nor real-time due to their reliance on scene-specific training, they both represent the strongest existing baselines for language-embedded reconstruction. 
In particular, LERF’s multiscale CLIP feature field avoids segmentation priors, making it non-object-centric and conceptually closest to OTAS.
We therefore include it despite the runtime mismatch, as it illustrates the trade-off between accurate but computationally expensive differential rendering approaches and our training-free, real-time alternative. 
We use a ROS bagfile of RoboNav\cite{Eder2023RoboNav}.
This allows for reproducible testing on real sensor data since the bagfile captures the full sensor and actuation context of the robot in representative environments.

Figure~\ref{fig:seg3d_robonav} shows language-embedded reconstructions of a challenging forest scene featuring dense vegetation and different ground types, such as grass, dirt and puddles.
LERF and Feature Splatting require highly accurate camera poses for reconstructing scenes with differential rendering.
Usually, Structure from Motion, like COLMAP \cite{schoenberger2016colmap}, is used for camera pose initialisation.
However, due to the cluttered, highly-textured scene, COLMAP, UKF Robot Localisation~\cite{moore2016generalized}, and VGGT~\cite{Wang2025VGGT} fail to provide poses with sufficient accuracy, see Figure~\ref{fig:seg3d_robonav} a).
Hence, camera poses are initialised using VGGT, scaled using metric depth estimation~\cite{Bhat2023Zoedepth}, and refined using Nerfacto \cite{nerfstudio}.
Even with pose refinement, Feature Splatting fails to properly reconstruct the ground. 
LERF correctly locates the grass-path itself and puddles (black circles) thanks to non-object-centric language grounding, Figure~\ref{fig:seg3d_robonav} b). 
However, it is computationally intensive with $\approx$40 minutes for this scene. 
OTAS shows a geometrically accurate reconstruction with detailed language similarity at $\approx$1.3 seconds, Figure~\ref{fig:seg3d_robonav} c). 
All runtime reports exclude pose initialisation and refinement.

\subsection{Ablations}

\textbf{Model Size and Inference Time (2D).} 
We provide multiple model configurations for different compute capabilities. 
Table \ref{tab:ablation_speed} presents their speed-accuracy trade-off on GPU and CPU.
Small and Large model configurations are outlined in Section~\ref{sec:experiments}.
\textit{No mask refinement} refers to normalising the similarity map $\mathcal{S}_{combined}$ to $[0,1]$ and binary thresholding.
No alignment with mask refinement represents $\mathcal{R}(I, \mathcal{U}_{bl}(\mathcal{S}(F_{text}, F_{vl})))$ and is equivalent to prompting SAM2.1 from CLIP similarity maps.
OTAS Small no refinement (line 3) significantly improves IoU and results in improved throughput, validating our token alignment strategy for feature regularisation.
Mask refinement further improves accuracy and adds $\approx$50\% to runtime on GPU.
\textit{OTAS Small} runs at real-time (assuming 10 fps).

\begin{table}[!b]
    \centering
    \caption{\textbf{Vision Backbones on RELLIS-3D.}
    Token alignment is evaluated across different frozen backbones using raw class labels as prompts and no mask refinement.
    OTAS achieves higher IoU than Grounded SAM-2 with an overall significantly lower parameter count, highlighting OTAS’ ability to regularise compact backbones into competitive open-vocabulary segmentation models without additional training.}
    \begin{tabular}{l"cr} \thickhline
        \textbf{Model Configuration} & mIoU(\%)$\uparrow$ & Param (M)$\downarrow$ \\ \hline
        Grounded SAM-2 & 45.11 & 396 \\  
        OTAS w. DINOv3 ViT-S/16 \cite{Simeoni2025Dinov3} & 48.44 & 107 \\
        OTAS w. C-RADIOv3-B \cite{Heinrich2024RADIO2} & 48.46 & 176 \\
        OTAS w. DINOv2 ViT-S/14 \cite{Oquab2023DINOv2} & 48.48 & 107 \\ \thickhline
    \end{tabular}
    \label{tab:ablation_foundation_model}
\end{table}

\textbf{Foundation Model Choice.}
Foundation model dependence is evaluated on RELLIS-3D \cite{Jiang2021Rellis}, a challenging off-road dataset with highly textured classes and semantically overlapping categories (e.g., “dirt,” “mud,” “puddle”).
Unlike purpose-trained methods that reach $\approx$75\% IoU on some classes \cite{Jiang2021Rellis}, open-vocabulary zero-shot approaches underperform due to ambiguous class boundaries.
We therefore use Rellis-3D as a close-to-real-world stress test across different vision backbones.
Table~\ref{tab:ablation_foundation_model} compares OTAS with Grounded SAM-2 and alternative foundation model choices for the visual encoder $\mathcal{E}_v$.
Alternative foundation models are DINOv3 \cite{Simeoni2025Dinov3}, a joint-embedding self-supervised model and successor to DINOv2, and AM-RADIO \cite{Heinrich2024RADIO2}, which is achieved by distilling multiple foundation models into a single backbone.
All experiments use raw class labels as prompts without tuning and a generic negative prompt of \textit{thing}.
Classes required for traversability assessment (i.e., dirt, water, asphalt, bush, mud, rubble) are evaluated with reported mean IoU (mIoU) over all classes equally weighted.
To isolate the performance of token alignment, mask refinement is deactivated for OTAS results.
Instead, similarity maps are thresholded at $0.8$.

All three foundation models combined with token alignment outperform Grounded SAM-2 despite using fewer parameters, showing that OTAS lifts frozen vision–language features into a more discriminative representation without training or fine-tuning additional segmentation heads.
However, performance when using the larger AM-RADIO (90M) backbone does not improve upon the significantly smaller DINO models (21M) when using the same CLIP image encoder for language-grounding (86M).

\textbf{Reduction and Clustering Methods.} 
Table \ref{tab:cluster-types} examines the choice of LVM ($\psi$) and clustering model ($C$) in a factorial experiment, using PCA (CPU and GPU), KPCA and ICA for LVM and k-Means (CPU and GPU), Gaussian Mixture Model (GMM) and HDBSCAN for clustering.
This comparison shows that k-Means clustering leads to the cleanest segmentation results, with PCA and Kernel-PCA being equally suitable for dimensionality reduction.
Density-based clustering (HDBSCAN) is a viable alternative if setting the number of clusters as a hyperparameter is not possible.

\begin{table}[]
    \centering
    \vspace{1mm}
    \caption{\textbf{Dimensionality Reduction and Clustering Algorithms.} 
    Score is the average of IoU, Fsc, Pre, and Rec of \textit{OTAS Small} with mask refinement on ORFD.}
    \begin{tabular}{l"cccc} \thickhline
        \textbf{Clustering} & PCA & KPCA & PCA (GPU) & ICA \\ \hline
        GMM          & 0.9381 & 0.9395 & 0.9390 & 0.9325 \\
        HDBSCAN      & 0.9394 & 0.9394 & 0.9394 & 0.9246 \\
        k-Means (GPU) & 0.9427 & 0.9427 & 0.9424 & 0.9312 \\
        k-Means      & \underline{0.9466} & \textbf{0.9467} & \textbf{0.9467} & 0.9363 \\
        \thickhline
    \end{tabular}
    \label{tab:cluster-types}
\end{table}

\begin{figure}[!h]
    \centering
    \includegraphics[width=0.99\linewidth]{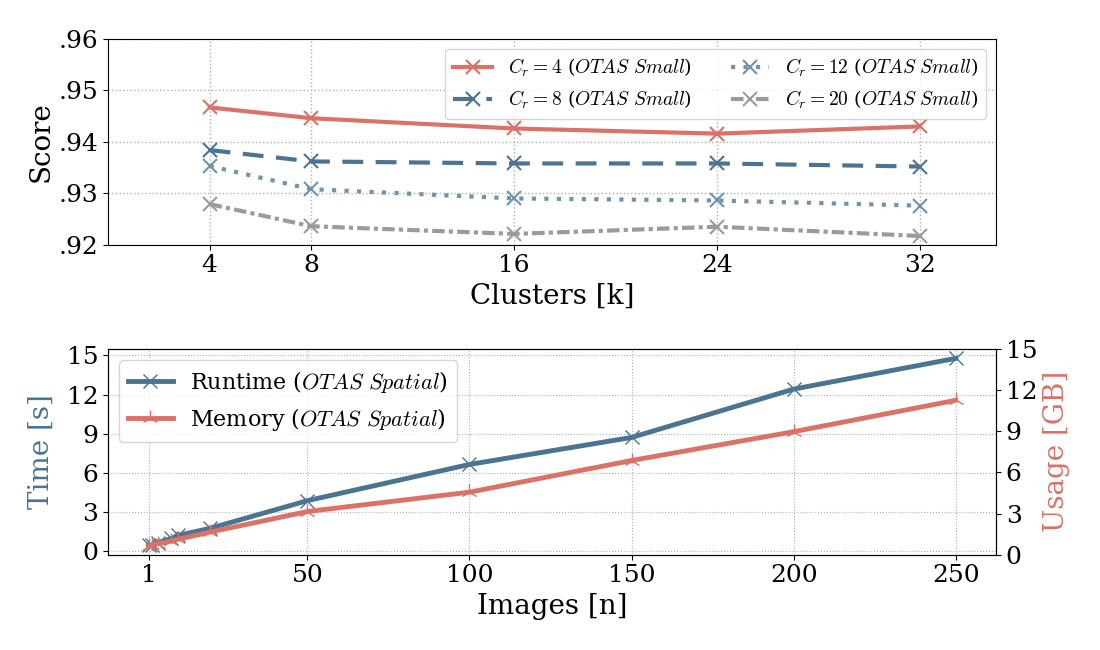}
    \caption{\textbf{Clusters, Components, Runtime and Memory.} Top presents the number of k-Means clusters ($k$) and number of components ($C_r$). Bottom shows runtime and memory usage.}
    \label{fig:cluster-numbers}
\end{figure}

\textbf{Number of Clusters and Components.} The top of Figure \ref{fig:cluster-numbers} shows an ablation of PCA components ($C_{r}$) and k-Means clusters ($k$) on a 20\% split of \ac{orfd}.
Results are truncated for length from a grid search over $k=[4;20]$, $C_r=[4;64]$ with marginal score difference between the best ($0.94$) and the worst-performing ($0.92$) combination.
Positive prompts are \textit{gravel, road, dirt} and negative prompts are \textit{sky, grass, forest}. 
The denoted score is an average of the IoU, F1 score, precision and recall. 
The highest score is achieved with $C_{r}=4$ and $k=4$.

\textbf{Model Scalability (3D).} 
Bottom of Figure~\ref{fig:cluster-numbers} shows the time requirements for reconstruction with \textit{OTAS Spatial} on TartanAir Seasonsforest Winter in blue, and the memory usage in orange. 
Both time and space complexity are comparably low to the state of the art and scale approximately linearly over the measured input view range.
At 10 views, both time and memory usage are marginally above 1 second and Gigabyte respectively.
Using 250 views takes 14.78 seconds and requires 11.26 Gigabytes of GPU memory, resulting in an average throughput of $\approx17$ fps.

\section{Conclusion}
\label{sec:conclusion}

This work addressed open-vocabulary segmentation in unstructured outdoor environments.
We introduce OTAS, an open-vocabulary segmentation method that aligns semantic tokens across single and multiple views to reconstruct a geometrically consistent feature field.
It aligns the output tokens of a pre-trained vision model to a language embedding by clustering semantically similar tokens through unsupervised learning and pooling.
OTAS is zero-shot, does not require scene-specific fine-tuning, and runs at up to $\approx$17 fps.
Results show a minor improvement over open-vocabulary and fine-tuned baselines on the ORFD dataset, a significant improvement over the state of the art on TartanAir, and robust applicability to real-world robotic tasks.
Scaling, runtime and backbone ablations confirm that OTAS is both efficient and backbone-agnostic, addressing concerns about model dependence and deployment trade-offs.
Future work will investigate employing our semantic maps for outdoor navigation, e.g., through costmap modification \cite{Qiu2024GeFF} or with learned policies \cite{Maheshwari2023PIAug}.


{\small
\bibliographystyle{IEEEtran}
\bibliography{root}

\begin{thebibliography}{10}
\providecommand{\url}[1]{#1}
\csname url@samestyle\endcsname
\providecommand{\newblock}{\relax}
\providecommand{\bibinfo}[2]{#2}
\providecommand{\BIBentrySTDinterwordspacing}{\spaceskip=0pt\relax}
\providecommand{\BIBentryALTinterwordstretchfactor}{4}
\providecommand{\BIBentryALTinterwordspacing}{\spaceskip=\fontdimen2\font plus
\BIBentryALTinterwordstretchfactor\fontdimen3\font minus \fontdimen4\font\relax}
\providecommand{\BIBforeignlanguage}[2]{{%
\expandafter\ifx\csname l@#1\endcsname\relax
\typeout{** WARNING: IEEEtran.bst: No hyphenation pattern has been}%
\typeout{** loaded for the language `#1'. Using the pattern for}%
\typeout{** the default language instead.}%
\else
\language=\csname l@#1\endcsname
\fi
#2}}
\providecommand{\BIBdecl}{\relax}
\BIBdecl

\bibitem{Huang2023VisualLanguageMaps}
C.~Huang, O.~Mees, A.~Zeng, and W.~Burgard, ``Visual language maps for robot navigation,'' in \emph{2023 IEEE International Conference on Robotics and Automation (ICRA)}, 2023, pp. 10\,608--10\,615.

\bibitem{Chen2023OpenVocabulary}
B.~Chen, F.~Xia, B.~Ichter, K.~Rao, K.~Gopalakrishnan, M.~S. Ryoo, A.~Stone, and D.~Kappler, ``Open-vocabulary queryable scene representations for real world planning,'' in \emph{2023 IEEE International Conference on Robotics and Automation (ICRA)}, 2023, pp. 11\,509--11\,522.

\bibitem{Gu2024conceptgraphs}
Q.~Gu, A.~Kuwajerwala, S.~Morin, K.~M. Jatavallabhula, B.~Sen, A.~Agarwal, C.~Rivera, W.~Paul, K.~Ellis, R.~Chellappa \emph{et~al.}, ``Conceptgraphs: Open-vocabulary 3d scene graphs for perception and planning,'' in \emph{2024 IEEE International Conference on Robotics and Automation (ICRA)}.\hskip 1em plus 0.5em minus 0.4em\relax IEEE, 2024, pp. 5021--5028.

\bibitem{Yamazaki2024OpenFusion}
K.~Yamazaki, T.~Hanyu, K.~Vo, T.~Pham, M.~Tran, G.~Doretto, A.~Nguyen, and N.~Le, ``Open-fusion: Real-time open-vocabulary 3d mapping and queryable scene representation,'' in \emph{2024 IEEE International Conference on Robotics and Automation (ICRA)}.\hskip 1em plus 0.5em minus 0.4em\relax IEEE, 2024, pp. 9411--9417.

\bibitem{Qiu2024Featuresplatting}
R.-Z. Qiu, G.~Yang, W.~Zeng, and X.~Wang, ``Language-driven physics-based scene synthesis and editing via feature splatting,'' in \emph{European Conference on Computer Vision (ECCV)}, 2024, pp. 368--383.

\bibitem{Kerr2023LERF}
J.~Kerr, C.~M. Kim, K.~Goldberg, A.~Kanazawa, and M.~Tancik, ``Lerf: Language embedded radiance fields,'' in \emph{2023 IEEE/CVF International Conference on Computer Vision (ICCV)}, 2023, pp. 19\,672--19\,682.

\bibitem{Cherti2023OpenCLIP}
M.~Cherti, R.~Beaumont, R.~Wightman, M.~Wortsman, G.~Ilharco, C.~Gordon, C.~Schuhmann, L.~Schmidt, and J.~Jitsev, ``Reproducible scaling laws for contrastive language-image learning,'' in \emph{Proceedings of the IEEE/CVF Conference on Computer Vision and Pattern Recognition (CVPR)}, June 2023, pp. 2818--2829.

\bibitem{Oquab2023DINOv2}
M.~Oquab, T.~Darcet, T.~Moutakanni, H.~V. Vo, M.~Szafraniec, V.~Khalidov, P.~Fernandez, D.~Haziza, F.~Massa, A.~El-Nouby, R.~Howes, P.-Y. Huang, H.~Xu, V.~Sharma, S.-W. Li, W.~Galuba, M.~Rabbat, M.~Assran, N.~Ballas, G.~Synnaeve, I.~Misra, H.~Jegou, J.~Mairal, P.~Labatut, A.~Joulin, and P.~Bojanowski, ``Dinov2: Learning robust visual features without supervision. \textit{arXiv preprint arXiv:2304.07193},'' 2023.

\bibitem{Radford2021CLIP}
A.~Radford, J.~W. Kim, C.~Hallacy, A.~Ramesh, G.~Goh, S.~Agarwal, G.~Sastry, A.~Askell, P.~Mishkin, J.~Clark, G.~Krueger, and I.~Sutskever, ``Learning transferable visual models from natural language supervision,'' in \emph{Proceedings of the 38th International Conference on Machine Learning}, ser. Proceedings of Machine Learning Research, M.~Meila and T.~Zhang, Eds., vol. 139.\hskip 1em plus 0.5em minus 0.4em\relax PMLR, 2021, pp. 8748--8763.

\bibitem{Kirillov2023SegmentAnything}
A.~Kirillov, E.~Mintun, N.~Ravi, H.~Mao, C.~Rolland, L.~Gustafson, T.~Xiao, S.~Whitehead, A.~C. Berg, W.-Y. Lo, P.~Dollar, and R.~Girshick, ``Segment anything,'' in \emph{Proceedings of the IEEE/CVF International Conference on Computer Vision (ICCV)}, October 2023, pp. 4015--4026.

\bibitem{Xueyan2023SEEM}
X.~Zou, J.~Yang, H.~Zhang, F.~Li, L.~Li, J.~Wang, L.~Wang, J.~Gao, and Y.~J. Lee, ``Segment everything everywhere all at once,'' in \emph{Advances in Neural Information Processing Systems}, A.~Oh, T.~Naumann, A.~Globerson, K.~Saenko, M.~Hardt, and S.~Levine, Eds., vol.~36.\hskip 1em plus 0.5em minus 0.4em\relax Curran Associates, Inc., 2023, pp. 19\,769--19\,782.

\bibitem{Ravi2024Sam2}
N.~Ravi, V.~Gabeur, Y.-T. Hu, R.~Hu, C.~Ryali, T.~Ma, H.~Khedr, R.~R{\"a}dle, C.~Rolland, L.~Gustafson, E.~Mintun, J.~Pan, K.~V. Alwala, N.~Carion, C.-Y. Wu, R.~Girshick, P.~Doll{\'a}r, and C.~Feichtenhofer, ``Sam 2: Segment anything in images and videos. \textit{arXiv preprint arXiv:2408.00714},'' 2024.

\bibitem{Chen2022ORFD_OFF_Net}
C.~Min, W.~Jiang, D.~Zhao, J.~Xu, L.~Xiao, Y.~Nie, and B.~Dai, ``Orfd: A dataset and benchmark for off-road freespace detection,'' in \emph{2022 International Conference on Robotics and Automation (ICRA)}, 2022, pp. 2532--2538.

\bibitem{Wang2020TartanAir}
W.~Wang, D.~Zhu, X.~Wang, Y.~Hu, Y.~Qiu, C.~Wang, Y.~Hu, A.~Kapoor, and S.~Scherer, ``Tartanair: A dataset to push the limits of visual slam,'' in \emph{2020 IEEE/RSJ International Conference on Intelligent Robots and Systems (IROS)}, 2020, pp. 4909--4916.

\bibitem{Nur2022CLIP-Fields}
N.~M.~M. Shafiullah, C.~Paxton, L.~Pinto, S.~Chintala, and A.~Szlam, ``Clip-fields: Weakly supervised semantic fields for robotic memory. \textit{arXiv preprint arXiv:2210.05663},'' 2022.

\bibitem{Qiu2024GeFF}
R.-Z. Qiu, Y.~Hu, Y.~Song, G.~Yang, Y.~Fu, J.~Ye, J.~Mu, R.~Yang, N.~Atanasov, S.~Scherer, and X.~Wang, ``Learning generalizable feature fields for mobile manipulation. \textit{arXiv preprint arXiv:2403.07563},'' 2024.

\bibitem{Yokoyama2024VLFM}
N.~Yokoyama, S.~Ha, D.~Batra, J.~Wang, and B.~Bucher, ``Vlfm: Vision-language frontier maps for zero-shot semantic navigation,'' in \emph{2024 IEEE International Conference on Robotics and Automation (ICRA)}, 2024, pp. 42--48.

\bibitem{li2023blip}
J.~Li, D.~Li, S.~Savarese, and S.~Hoi, ``Blip-2: Bootstrapping language-image pre-training with frozen image encoders and large language models,'' in \emph{International conference on machine learning}.\hskip 1em plus 0.5em minus 0.4em\relax PMLR, 2023, pp. 19\,730--19\,742.

\bibitem{Peng2023OpenScene}
S.~Peng, K.~Genova, C.~{\textquotedblleft}. Jiang, A.~Tagliasacchi, M.~Pollefeys, and T.~Funkhouser, ``Openscene: 3d scene understanding with open vocabularies,'' in \emph{Proceedings of the IEEE/CVF Conference on Computer Vision and Pattern Recognition (CVPR)}, June 2023, pp. 815--824.

\bibitem{Zhang2025QuantifyingLimitsSAM}
Y.~Zhang, N.~Konz, K.~Kramer, and M.~A. Mazurowski, ``Quantifying the limits of segmentation foundation models: Modeling challenges in segmenting tree-like and low-contrast objects. \textit{arXiv preprint arXiv:2412.04243},'' 2025.

\bibitem{Zhou2021MaskCLIP}
C.~Zhou, C.~C. Loy, and B.~Dai, ``Extract free dense labels from clip,'' in \emph{Computer Vision -- ECCV 2022}, S.~Avidan, G.~Brostow, M.~Ciss{\'e}, G.~M. Farinella, and T.~Hassner, Eds.\hskip 1em plus 0.5em minus 0.4em\relax Cham: Springer Nature Switzerland, 2022, pp. 696--712.

\bibitem{Jiang2021Rellis}
P.~Jiang, P.~Osteen, M.~Wigness, and S.~Saripalli, ``Rellis-3d dataset: Data, benchmarks and analysis,'' in \emph{2021 IEEE International Conference on Robotics and Automation (ICRA)}, 2021, pp. 1110--1116.

\bibitem{Alama2025RayFronts}
O.~Alama, A.~Bhattacharya, H.~He, S.~Kim, Y.~Qiu, W.~Wang, C.~Ho, N.~Keetha, and S.~Scherer, ``Rayfronts: Open-set semantic ray frontiers for online scene understanding and exploration. \textit{arXiv preprint arXiv:2504.06994},'' 2025.

\bibitem{Min2024AutonomousUnstructuredReview}
C.~Min, S.~Si, X.~Wang, H.~Xue, W.~Jiang, Y.~Liu, J.~Wang, Q.~Zhu, Q.~Zhu, L.~Luo, F.~Kong, J.~Miao, X.~Cai, S.~An, W.~Li, J.~Mei, T.~Sun, H.~Zhai, Q.~Liu, F.~Zhao, L.~Chen, S.~Wang, E.~Shang, L.~Shang, K.~Zhao, F.~Li, H.~Fu, L.~Jin, J.~Zhao, F.~Mao, Z.~Xiao, C.~Li, B.~Dai, D.~Zhao, L.~Xiao, Y.~Nie, Y.~Hu, and X.~Li, ``Autonomous driving in unstructured environments: How far have we come?, radiological, and nuclear disaster response. \textit{arXiv preprint arXiv:2410.07701},'' 2024.

\bibitem{Eder2023RoboNav}
M.~Eder, R.~Prinz, F.~Schöggl, and G.~Steinbauer-Wagner, ``Traversability analysis for off-road environments using locomotion experiments and earth observation data,'' \emph{Robotics and Autonomous Systems}, vol. 168, p. 104494, 2023.

\bibitem{Darcet2023VitRegisters}
T.~Darcet, M.~Oquab, J.~Mairal, and P.~Bojanowski, ``Vision transformers need registers. \textit{arXiv preprint arXiv:2309.16588},'' 2023.

\bibitem{Jiahang2024RoadFormer}
J.~Li, Y.~Zhang, P.~Yun, G.~Zhou, Q.~Chen, and R.~Fan, ``Roadformer: Duplex transformer for rgb-normal semantic road scene parsing,'' \emph{IEEE Transactions on Intelligent Vehicles}, vol.~9, no.~7, pp. 5163--5172, 2024.

\bibitem{Sun2019RTFNet}
Y.~Sun, W.~Zuo, and M.~Liu, ``Rtfnet: Rgb-thermal fusion network for semantic segmentation of urban scenes,'' \emph{IEEE Robotics and Automation Letters}, vol.~4, no.~3, pp. 2576--2583, 2019.

\bibitem{Hongliang2023M2F2-Net}
H.~Ye, J.~Mei, and Y.~Hu, ``M2f2-net: Multi-modal feature fusion for unstructured off-road freespace detection,'' in \emph{2023 IEEE Intelligent Vehicles Symposium (IV)}, 2023, pp. 1--7.

\bibitem{Lv2024NaifNet}
Y.~Lv, Z.~Liu, G.~Li, and X.~Chang, ``Noise-aware intermediary fusion network for off-road freespace detection,'' \emph{IEEE Transactions on Intelligent Vehicles}, pp. 1--11, 2024.

\bibitem{Ren2024GroundedSAM}
T.~Ren, S.~Liu, A.~Zeng, J.~Lin, K.~Li, H.~Cao, J.~Chen, X.~Huang, Y.~Chen, F.~Yan, Z.~Zeng, H.~Zhang, F.~Li, J.~Yang, H.~Li, Q.~Jiang, and L.~Zhang, ``Grounded sam: Assembling open-world models for diverse visual tasks. \textit{arXiv preprint arXiv:2401.14159},'' 2024.

\bibitem{IDEAResearchGroundedSAM2}
IDEA-Research, ``Grounded-sam-2: Ground and track anything in videos with grounding dino, florence-2, and sam 2,'' \url{https://github.com/IDEA-Research/Grounded-SAM-2}, 2025, accessed: 2025-04-30.

\bibitem{schoenberger2016colmap}
J.~L. Sch\"{o}nberger and J.-M. Frahm, ``Structure-from-motion revisited,'' in \emph{Conference on Computer Vision and Pattern Recognition (CVPR)}, 2016, pp. 4104--4113.

\bibitem{moore2016generalized}
T.~Moore and D.~Stouch, ``A generalized extended kalman filter implementation for the robot operating system,'' in \emph{Intelligent Autonomous Systems 13: Proceedings of the 13th International Conference IAS-13}.\hskip 1em plus 0.5em minus 0.4em\relax Springer, 2016, pp. 335--348.

\bibitem{Wang2025VGGT}
J.~Wang, M.~Chen, N.~Karaev, A.~Vedaldi, C.~Rupprecht, and D.~Novotny, ``Vggt: Visual geometry grounded transformer,'' in \emph{Proceedings of the IEEE/CVF Conference on Computer Vision and Pattern Recognition}, 2025.

\bibitem{Bhat2023Zoedepth}
S.~F. Bhat, R.~Birkl, D.~Wofk, P.~Wonka, and M.~Müller, ``Zoedepth: Zero-shot transfer by combining relative and metric depth. . \textit{arXiv preprint arXiv:2302.12288},'' 2023.

\bibitem{nerfstudio}
M.~Tancik, E.~Weber, E.~Ng, R.~Li, B.~Yi, J.~Kerr, T.~Wang, A.~Kristoffersen, J.~Austin, K.~Salahi, A.~Ahuja, D.~McAllister, and A.~Kanazawa, ``Nerfstudio: A modular framework for neural radiance field development,'' in \emph{ACM SIGGRAPH 2023 Conference Proceedings}, ser. SIGGRAPH '23, 2023.

\bibitem{Simeoni2025Dinov3}
O.~Sim{\'e}oni, H.~V. Vo, M.~Seitzer, F.~Baldassarre, M.~Oquab, C.~Jose, V.~Khalidov, M.~Szafraniec, S.~Yi, M.~Ramamonjisoa, F.~Massa, D.~Haziza, L.~Wehrstedt, J.~Wang, T.~Darcet, T.~Moutakanni, L.~Sentana, C.~Roberts, A.~Vedaldi, J.~Tolan, J.~Brandt, C.~Couprie, J.~Mairal, H.~J{\'e}gou, P.~Labatut, and P.~Bojanowski, ``{DINOv3}. \textit{arXiv preprint arXiv:2508.10104},'' 2025.

\bibitem{Heinrich2024RADIO2}
G.~Heinrich, M.~Ranzinger, Hongxu, Yin, Y.~Lu, J.~Kautz, A.~Tao, B.~Catanzaro, and P.~Molchanov, ``Radiov2.5: Improved baselines for agglomerative vision foundation models. \textit{arXiv preprint arXiv:2412.07679},'' 2024.

\bibitem{Maheshwari2023PIAug}
P.~Maheshwari, W.~Wang, S.~Triest, M.~Sivaprakasam, S.~Aich, J.~G.~R. III, J.~M. Gregory, and S.~Scherer, ``Piaug -- physics informed augmentation for learning vehicle dynamics for off-road navigation. \textit{arXiv preprint arXiv:2311.00815},'' 2023.

\end{thebibliography}
}

\clearpage

\clearpage
\twocolumn[
\begin{center}
    {\Large \bfseries Supplementary Material for\\OTAS: Open-vocabulary Token Alignment for Outdoor Segmentation\\[1ex]}
\end{center}
\vspace{1em}
]

\section{Supplementary Feature Reconstruction Results}

\begin{figure*}[!h]
    \centering
    \includegraphics[width=0.87\linewidth]{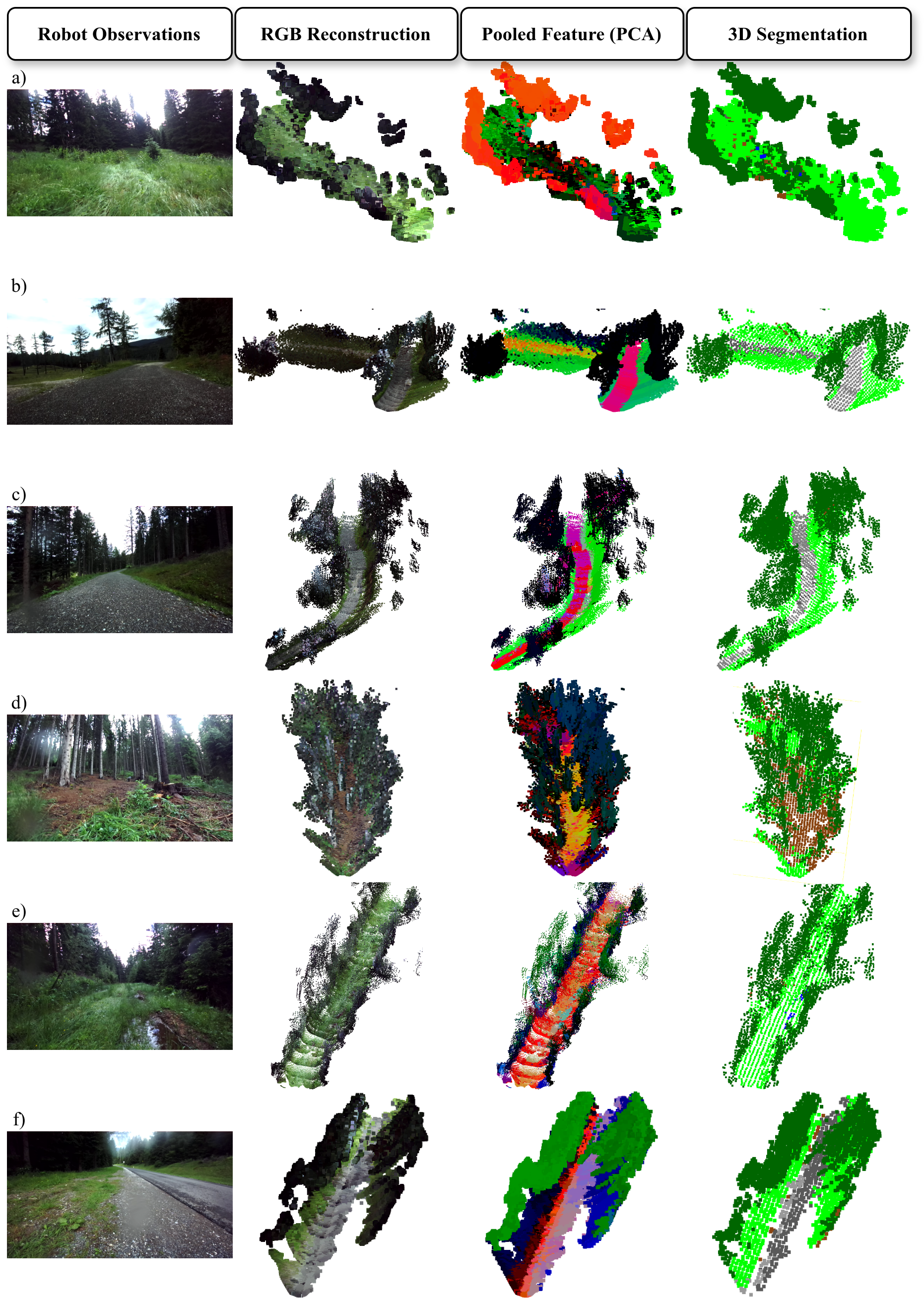}
    \caption{\textbf{Alpine 3D Segmentation.} Left-to-right: Input image (of $n$), RGB point cloud $P_{global}$, PCA over $P_{Semantic}$ and segmentation over $\hat{P}_{Semantic}$. Scenes are a) open field with shrubbery, b) cross-road between a road and field track, c) forest road, d) steep area including trees, bushes and duff, e) grass path with puddles, and f) asphalt street. Semantic classes \textit{trees}, \textit{shrubs}, and \textit{bushes} are dark green, \textit{grass} is light green, \textit{road} and \textit{gravel} are dark and light grey, \textit{duff} is brown, and \textit{puddles} are blue.}
    \label{fig:appendix_robonav}
\end{figure*}

Figure~\ref{fig:appendix_robonav} presents \textit{OTAS Spatial} reconstructions on RoboNav~\cite{Eder2023RoboNav}.  
The leftmost column shows an image of the scenes.
The second column to the left shows  the RGB reconstruction of $P_{global}$.  
Language-grounded semantic information is visualised using PCA over $P_{Semantic}$ in the third column, and a 3D segmentation using common labels in outdoor robotics (rightmost column).  
Segmentations are obtained by thresholding the similarity between $\hat{P}_{Semantic}$ and a set of text prompts for each scene.  
\textit{Grass} is depicted in light green, \textit{trees} and \textit{shrubbery} in dark green, the \textit{duff layer} (comprising dead leaves and small twigs) in brown, \textit{gravel} in light grey, \textit{road} in dark grey, and \textit{puddle} and \textit{water} in blue.

\textbf{Figure~\ref{fig:appendix_robonav}a)} depicts a clearing in a forest.  
The primary challenge in this scene is to differentiate between tall grass, vegetation flattened by repeated vehicular traffic, and shrubbery located in the centre of the scene.  
The geometric reconstruction shows high visual similarity of tall grass and shrubbery, yet they have distinct implications for traversability.  
In the PCA visualisation, trees and shrubbery are indicated as semantically similar (red), and are clearly separated from grass (green).  
At the boundaries between semantic classes, the dark black regions indicate ambiguous semantic associations between tall grass and shrubbery in the raw feature reconstruction.  
Nevertheless, when features are queried using open-vocabulary prompts, OTAS successfully segments the narrow grass path for traversing the shrubbery.

\textbf{Figure~\ref{fig:appendix_robonav}b)} illustrates a crossroad of a gravel road and a field track.  
The primary challenge is in correctly identifying both roads for obtaining information about traversability.  
The road on the right is a forest road composed of gravel, while the left path is a field track.  
The PCA visualisation shows that the road and the path are semantically similar (pink and orange) in OTAS' feature reconstruction.
Through the open-vocabulary prompt \textit{"road"}, OTAS correctly identifies both as traversable areas (grey).

\textbf{Figure~\ref{fig:appendix_robonav}c)} presents a forest road under varying lighting conditions, a common scenario in outdoor robotics.  
OTAS clearly distinguishes the road (grey) from vegetation (dark green) and grass (light green).

\textbf{Figure~\ref{fig:appendix_robonav}d)} shows a steep, highly unstructured area containing grass, trees, bushes and duff.  
The primary challenge in this scene is to correctly distinguish between different ground types, namely tall grass, shrubbery, tree stumps, and duff.  
This is particularly challenging as these ground types blend into each other, leading to fuzzy semantic boundaries. 
Furthermore, the ground is cluttered.  
The PCA reconstruction demonstrates that in the language-grounded embeddings, the ground types can be clearly distinguished from each other.  
Shrubbery and tree stumps are dark green, tall grass is red, and duff is yellow.  
The PCA visualisation also illustrates how these ground types mix with each other in the raw language-grounded embeddings, depicting how they blend into each other in reality. 
When prompted, the resulting segmentation distinguishes between \textit{grass} (light green), \textit{shrubs} and \textit{trees} (dark green), and \textit{duff} (brown).

\textbf{Figure~\ref{fig:appendix_robonav}e)} shows a grass path with puddles.  
The primary challenge is to correctly distinguish the wet ground and puddles from the traversable grass.  
In the PCA visualisation, the \textit{puddles} (blue) are clearly separated from the \textit{grass} (red).  
However, in the prompt-based segmentation, the \textit{puddles} (blue) are not as clearly defined, indicating the need for a smaller voxel size.

\textbf{Figure~\ref{fig:appendix_robonav}f)} shows an asphalt road going downhill.  
The street smoothly transitions to the gravel strip on the left, which itself blends into grass and vegetation.  
This scene presents very unclear semantic boundaries. 
OTAS' PCA visualisation shows that the ground types are clearly distinguishable in the semantics. 
Furthermore, the blending of ground types is also apparent in the raw language-grounded embeddings.  
\textit{Asphalt} (purple) mixes with \textit{gravel} (red) and \textit{grass} (dark red).  
In the prompt-based segmentation, OTAS determines clear boundaries between these ground types.

\clearpage

\section{Supplementary Experimental Detail}

\subsection{Datasets}

\textbf{Off-Road Freespace Detection (ORFD)~\cite{Chen2022ORFD_OFF_Net}}
ORFD experiments use $k=4$ clusters and $C_{r}=4$ components, with positive prompts \textit{"gravel", "road", "dirt"} and negative prompts \textit{"sky", "grass", "forest"}. For OTAS without mask refinement, a similarity threshold of $0.5$ is considered a positive label.
SEEM~\cite{Xueyan2023SEEM} and Grounded-SAM~\cite{Ren2024GroundedSAM} use the same positive prompts as OTAS. 
Negative prompts are not supported by them.
Grounded-SAM 2~\cite{IDEAResearchGroundedSAM2} neither supports negative nor multiple prompt input. It also does not support whitespaces in prompts and requires prompts to end with a period. To ensure a best-case scenario for Grounded-SAM 2 and subsequent fair comparison, all positive prompts used for the other methods where tested, of which \textit{"road."} achieved the best results across ORFD experiments.

\textbf{TartanAir~\cite{Wang2020TartanAir}}
The TartanAir experiments use $k=30$ clusters and $C_{r}=30$ components, with positive prompts \textit{"tree", "bush", "vegetation"}, and negative prompts \textit{"sky", "stone", "object"}, again with a similarity threshold of $0.5$. 
Invalid, infinite, and depth values over $150$ metres are discarded.
For sequences too large to fit into 16GB of GPU VRAM, only every third image is used for reconstruction. This was necessary for OpenFusion on all sequences, and for \textit{OTAS Spatial} on Amusement and Gascola.
The baselines OpenFusion~\cite{Yamazaki2024OpenFusion} and ConceptGraphs~\cite{Gu2024conceptgraphs} are configured to output point cloud reconstructions with per-point CLIP~\cite{Radford2021CLIP} features.
These point clouds are prompted identically to the OTAS reconstructions.
For ConceptGraphs, we used the configuration of the original paper based on SAM.

\textbf{RoboNav~\cite{Eder2023RoboNav}}
COLMAP~\cite{schoenberger2016colmap} with the unaltered camera stream and default matcher parameters failed to provide camera poses with sufficient accuracy for obtaining reconstructions.
Hence, input images were pre-processed with a sharpening kernel, brightness and contrast adjustment, and fast non-local means denoising.
This enabled pose initialisation using COLMAP's sequential matcher with relaxed parameters: overlap ($10$), quadratic overlap disabled, loop detection every $10$ frames, a reduced loop detection window ($50$ images), fewer nearest neighbours ($5$), and a reduced number of checks ($256$).
UKF Robot localisation fuses wheel odometry with GNSS using an Unscented Kalman Filter. 

\textbf{RELLIS-3D~\cite{Jiang2021Rellis}}
All models are tested with a native $64\times64$ patch grid and without mask refinement to isolate foundation model performance.
Input images are resized accordingly to $1024\times1024$ to achieve the shared feature resolution of $d=64$ without interpolation for AM-RADIO and DINOv3.
To adhere to DINOv2's recommended maximum input resolution, DINOv2 extracts a $16\times16$ feature grid with bilinear feature interpolation.
Results are all obtained with $k=24$ clusters and $C_r=24$ components, set empirically as a good starting point without dataset-specific tuning.
All prompts are class names in RELLIS-3D's included ontology file.
For Grounded-SAM 2 results, prompt names have a period appended, to adhere to the required prompt format.

\subsection{Feature Reconstruction Baselines} 
VGGT~\cite{Wang2025VGGT} does not provide metric depth.
Consequently images are scaled by the inverse ratio between VGGT's and metric depth.
Scaled camera extrinsics and depth maps are then used for reconstruction.
Areas depicting the sky (obtained using single-view segmentation on \textit{OTAS Small} with positive \textit{"sky", "clouds"} and negative \textit{"ground", "object"} prompts) and overexposed areas (obtained using a threshold of $0.75$) are excluded from scale estimation.
For \textit{OTAS Spatial}, these areas and depth values over $15$ metres are also excluded from the reconstruction.

Nerfacto~\cite{nerfstudio}, Feature Splatting~\cite{Qiu2024Featuresplatting}, and LERF~\cite{Kerr2023LERF} are trained using default settings and for the default number of iterations.
\textit{OTAS Spatial} uses $k=12$ clusters and $C_{r}=24$ components.
Qualitative results use the prompts and segmentation thresholds \textit{"grass"} ($0.5$), \textit{"gravel"} ($0.8$), \textit{"road"} ($0.65$), \textit{"tree"} ($0.5$), \textit{"shrubbery"} ($0.6$), \textit{"tree stump"} ($0.8$), \textit{"duff layer"} ($0.7$), and \textit{"water"} ($0.9$).
Each similarity query also includes the positive prompt of \textit{"ground"} and a negative prompt of \textit{"object"} to combat noisy depth predictions.

Visualised point clouds have their outliers removed using radius outlier rejection (removing points with fewer than $9$ neighbours within a $4$ metre radius) followed by statistical outlier rejection (removing points with distances to their $30$ nearest neighbours larger than $1.8$ standard deviations from the mean distance).
Points with more than $0.75$ brightness are removed as they are likely overexposed.
For semantic visualisation, the point cloud has all points further than $0.5$ metres from the geometric point cloud removed.
These steps are exclusively cosmetic to improve the clarity of the visualisations.

\end{document}